\documentclass[runningheads]{llncs}
\usepackage{graphicx}

\usepackage{tikz}
\usepackage{comment}
\usepackage{amsmath,amssymb} 
\usepackage{color}
\usepackage{orcidlink}

\usepackage[accsupp]{axessibility}  

\usepackage{dsfont}
\usepackage{subcaption}
\usepackage{multirow}
\usepackage{makecell}
\usepackage{hyperref}
\usepackage{marvosym}
\usepackage{pifont}

\begin{document}
\pagestyle{headings}
\mainmatter
\def\ECCVSubNumber{160}  

\title{Temporal Lift Pooling for Continuous Sign Language Recognition} 

\titlerunning{Temporal Lift Pooling for Continuous Sign Language Recognition}
%
\author{Lianyu Hu\inst{1}\orcidlink{0000-0003-2470-8110} \and
Liqing Gao\inst{1}\orcidlink{0000-0003-4518-2154} \and
Zekang Liu\inst{1}\orcidlink{0000-0002-5003-1900}\and
Wei Feng\textsuperscript{\Letter}\inst{1}\orcidlink{0000-0003-3809-1086}}
\authorrunning{Lianyu et al.}
%
\institute{Tianjin University, Tianjin 300350, China
\email{\{hly2021,lqgao,lzk100953\}@tju.edu.cn;wfeng@ieee.org}}
\maketitle

\begin{abstract}
Pooling methods are necessities for modern neural networks for increasing receptive fields and lowering down computational costs. However, commonly used hand-crafted pooling approaches, e.g., max pooling and average pooling, may not well preserve discriminative features. While many researchers have elaborately designed various pooling variants in spatial domain to handle these limitations with much progress, the temporal aspect is rarely visited where directly applying hand-crafted methods or these specialized spatial variants may not be optimal. In this paper, we derive temporal lift pooling (TLP) from the Lifting Scheme in signal processing to intelligently downsample features of different temporal hierarchies. The Lifting Scheme factorizes input signals into various sub-bands with different frequency, which can be viewed as different temporal movement patterns. Our TLP is a three-stage procedure, which performs signal decomposition, component weighting and information fusion to generate a refined downsized feature map. We select a typical temporal task with long sequences, i.e. continuous sign language recognition (CSLR), as our testbed to verify the effectiveness of TLP. Experiments on two large-scale datasets show TLP outperforms hand-crafted methods and specialized spatial variants by a large margin (1.5\%) with similar computational overhead. As a robust feature extractor, TLP exhibits great generalizability upon multiple backbones on various datasets and achieves new state-of-the-art results on two large-scale CSLR datasets. Visualizations further demonstrate the mechanism of TLP in correcting gloss borders. Code is released\footnote{https://github.com/hulianyuyy/Temporal-Lift-Pooling}.   
\end{abstract}

\section{Introduction}
\begin{figure}[t]
\centering
\includegraphics[width=0.95\columnwidth]{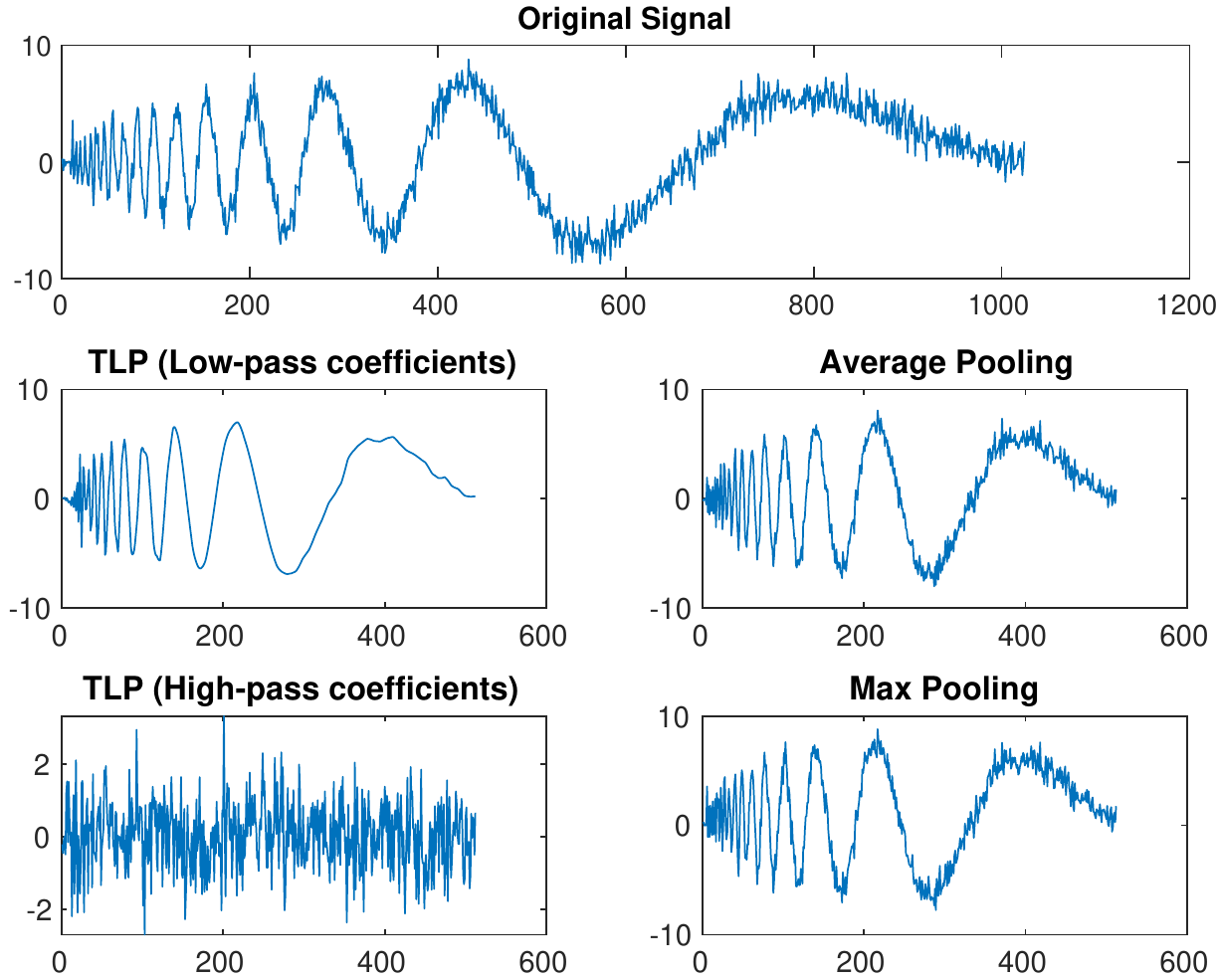} 
\caption{Effects of temporal lift pooling (TLP) and hand-crafted pooling methods. TLP clearly decomposes input signal into various temporal patterns while hand-crafted pooling methods can't well distinguish noise from body movements.}
\label{fig1}
\end{figure}
Sign language is one of the most commonly used communication tools for the deaf people. However, mastering this language is difficult for the hearing people, which forms an obstacle for communication between two groups. To handle this problem, continuous sign language recognition (CSLR) aims to translate sign videos into corresponding gloss sentences, which is feasible to bridge this gap.

Pooling methods are necessities for modern neural networks (NNs) for increasing receptive fields and generating discriminative representations. Several simple pooling methods, like max pooling and average pooling, are broadly employed in various domains~\cite{boureau2010theoretical,lecun1989handwritten,yu2014mixed} for their remarkable generalizability. While effective and efficient, simply using these hand-crafted methods may not fully consider local structures and optimally preserve features of different hierarchies. For the spatial domain, many researchers~\cite{yu2014mixed,gulcehre2014learned,zhai2017s3pool,saeedan2018detail,stergiou2021refining,zeiler2013stochastic,gao2019lip} have realized the limitations of hand-crafted pooling and elaborately designed many downsampling approaches for better preserving details. However, the temporal aspect is rarely explored where directly applying hand-crafted methods or these spatially specialized variants may not fit the temporal pattern well.

Our method is inspired by the Lifting Scheme~\cite{sweldens1998lifting} from signal processing, which is commonly used in information compression~\cite{pesquet2001three}, reconstruction~\cite{dogiwal2014efficient} and denoising~\cite{wu2004adaptive}. The Lifting Scheme decomposes an input signal into various sub-bands with downscaled sizes of different frequencies, which is ideal for joint time-frequency analysis. Applying the idea of Lifting Scheme, we present temporal lift pooling (TLP) to factorize inputs into major and adjunctive movements and integrate them into a downsized refined representation. As shown in Fig.~\ref{fig1}, the low-pass coefficients generated by TLP smoothly restore original low-frequency signals, which can be viewed as body movement patterns. The high-pass coefficients extract high-frequency components from inputs that represent detailed dynamics. In contrast, hand-crafted max pooling and average pooling fail to deal with mixed inputs and even amplify extremes or lose details sometimes.

TLP is consisted of three stages: lifting process, component weighting and fusion, which step by step decomposes input signal and reweights its components for a unified output. As a plug-and-play tool, TLP is implemented with tiny convolutional neural networks with only additional 0.4\% computational costs. As an effective downsampling unit, it exhibits excellent generalizability upon multiple backbones on various datasets. By only replacing two pooling locations with TLP, a significant 1.5\% performance boost is witnessed, which largely surpasses the hand-crafted methods and spatial pooling variants on CSLR. Besides, TLP achieves new state-of-the-art results on two large-scale CSLR datasets. Visualizations present the effects of TLP to correct gloss borders on accurate recognition.

\section{Related Work}
\subsection{Continuous Sign Language Recognition}
Earlier methods~\cite{gao2004chinese,han2009modelling,freeman1995orientation,koller2015continuous} in CSLR always employ hand-crafted features or HMM-based systems~\cite{koller2016deepsign,koller2017re} to perform temporal modeling and translate sentences step by step. HMM-hybrid methods~\cite{koller2016deepsign,koller2017re} typically first employ a feature extractor for representative features and then adopt an HMM for long-term temporal modeling. The success of convolutional neural networks (CNNs) and recurrent neural networks (RNNs) bring huge progress for CSLR. CTC loss~\cite{graves2006connectionist} provides a new perspective to align target sentences with input frames which is broadly used by recent CSLR approaches~\cite{pu2019iterative,pu2020boosting,cheng2020fully,cui2019deep,niu2020stochastic,Min_2021_ICCV}. They first rely on a feature extractor, i.e. 3D or 2D\&1D CNN hybrids, to extract frame-wise features, and then adopt a LSTM module for capturing long-term temporal dependencies. However, several methods~\cite{pu2019iterative,cui2019deep} found in such conditions the feature extractor is not well trained. Some recent approaches present the iterative training strategy to relieve this problem, but consume much more computations and multiple training stages. More recent studies~\cite{Min_2021_ICCV,cheng2020fully,pu2020boosting} try to directly enhance the feature extractor by adding alignment losses~\cite{Min_2021_ICCV} or adopt pseudo labels for supervision to tackle this issue~\cite{cheng2020fully}. 

\subsection{Pooling Methods}
Pooling has been commonly used in modern NNs for discriminative representations and reducing computational costs since Neocognitron~\cite{fukushima1982neocognitron}. Previous studies mainly focus on pooling in the spatial domain but rarely explore the temporal side. Max pooling and average pooling are two commonly used flexible hand-crafted methods in various tasks which could be dated back to LeNet~\cite{lecun1989handwritten} periods. Boureau et al.~\cite{boureau2010theoretical} prove max pooling can preserve more discriminative features than average pooling in terms of probability. Apart from these simple hand-crafted methods, various pooling variants have been proposed to better preserve details while maintaining efficiency.  $L_p$ pooling~\cite{gulcehre2014learned} introduces $L_p$ norm to activate and normalize outputs, which can be viewed as a continuum between max and average pooling. Mixed pooling~\cite{yu2014mixed} tries to integrate the characteristics of max pooling and average pooling by a learned coefficient for better performance. Stochastic pooling~\cite{zeiler2013stochastic} presents a multinomial sampling algorithm to select output values in the sampling window. S3Pool~\cite{zhai2017s3pool} attempts to introduce regulation in rows and columns, which can be viewed as some kind of data augmentation. Detail-preserving pooling (DPP)~\cite{saeedan2018detail} aims to preserve details in 2D grids by selecting discriminative responses. LIP~\cite{gao2019lip} formulates existing pooling methods under a general framework and designs a tiny convolutional network to generate local importance for values in a sampling window. Softpool~\cite{stergiou2021refining} employs the softmax function to measure the contribution of values and adopts the normalized outputs as downscaled contents. LiftPool~\cite{zhao2021liftpool} introduces Lifting Scheme to design both downsampling and upsampling variants. However, it mainly tackles spatial tasks and doesn't consider temporal patterns. Besides, it fails to further measure the contribution of different components in sub-bands and doesn't deal with hierarchical features. Especially, all these approaches focus on the spatial aspect but don't explore the temporal side, while not all of them (e.g., DPP~\cite{saeedan2018detail} and S3Pool~\cite{zhai2017s3pool}) are directly applicable for temporal tasks. Besides, directly applying hand-crafted pooling methods or these specialized spatial variants may not be optimal for temporal modeling. As shown in the experiments, our TLP surpasses all these counterparts by a large margin.

\section{Methods}
\subsection{Overview}

\begin{figure}[t]
\centering
\includegraphics[width=\columnwidth]{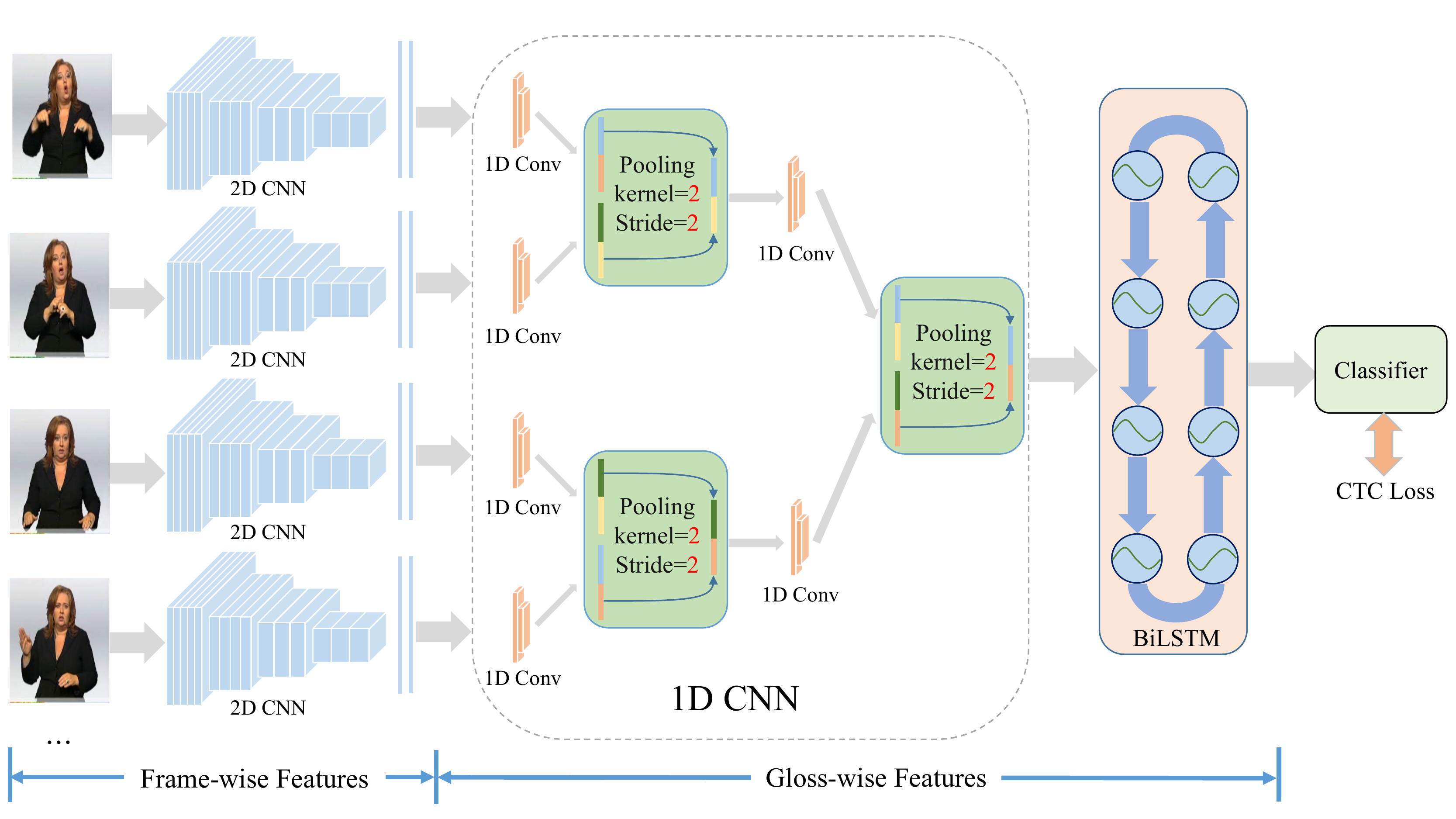} 
\caption{Overview of recent CSLR methods. They first employ a common 2D CNN to extract frame-wise features, and then adopt a 1D CNN to perform short-term temporal modeling. A two-layer BiLSTM is used to capture long-term temporal dependencies, followed by a classifier to predict sentences. Especially, two pooling layers are adopted in 1D CNN for shortened discriminative representation. Practically, max pooling with kernel size of 2 and stride of 2 is used as default. We replace them with TLP to intelligently preserve discriminative features.}
\label{fig2}
\end{figure}

As shown in Fig.~\ref{fig2}, recent CSLR methods~\cite{pu2019iterative,pu2020boosting,cheng2020fully,cui2019deep,niu2020stochastic,Min_2021_ICCV} usually first employ a common 2D CNN to extract frame-wise features, and then deploy a 1D CNN consisted of a sequence of 1D Conv and pooling methods to model short-term temporal dependencies, followed by a two-layer BiLSTM and classifier for sentence prediction. Especially, two pooling layers are adopted in the 1D CNN to squeeze input length for downsampled discriminative representations to predict sentences. Practically, max pooling with kernel size of 2 and stride of 2 is used as default. As the downsampling process is intrinsically lossy, it's necessary to consider which information to be kept for subsequent sentence prediction. Inappropriate downsampling may lead to beneficial information loss and movement pattern deformation, thus affecting recognition performance. In this paper, we refer to Lifting Scheme~\cite{sweldens1998lifting} originated from signal processing to handle this issue and derive an efficient and effective pooling method.
\subsection{Temporal Lift Pooling}

Pooling methods are necessities for reducing computational costs and obtaining discriminative representations for temporal tasks with long input sequences, e.g., CSLR. Commonly used hand-crafted pooling methods may not well consider local patterns and don't optimally preserve critical representations. We derive temporal lift pooing (TLP) from the Lifting Scheme to exploit temporal correlations in signals to build a downsized approximation. 

\begin{figure}[t]
\centering
\includegraphics[width=0.8\columnwidth]{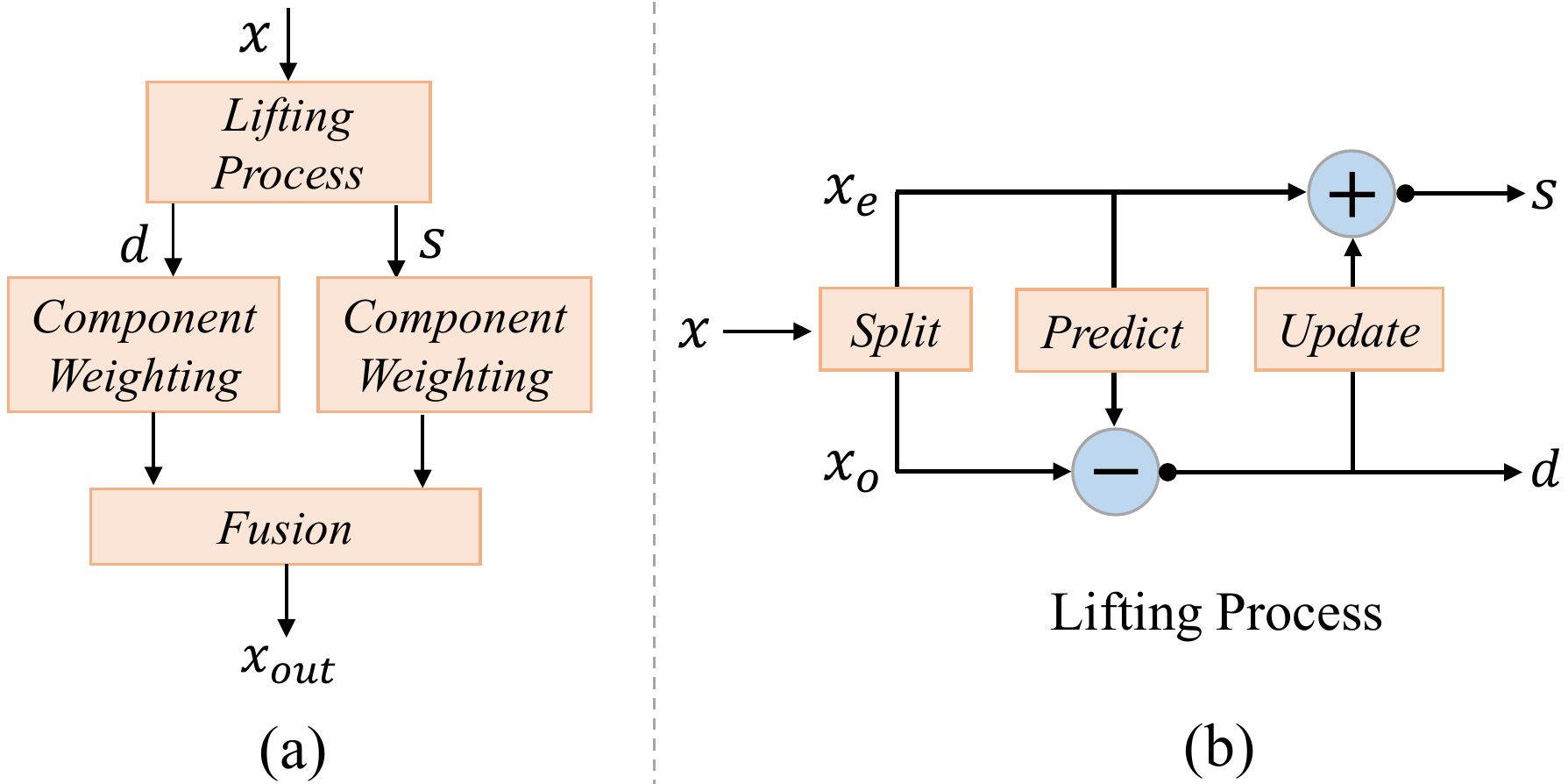} 
\caption{(a) Overview for temporal lift pooling of three stages: lifting process, component weighting and fusion. (b) Lifting process. $x$ is split into $x_e$ and $x_o$, where the predictor and updater generate an approximation $s$ and a difference signal $d$.}
\label{fig3}
\end{figure}

As shown in Fig.~\ref{fig3}(a), our TLP is composed of three stages, i.e., lifting process, component weighting and fusion. We will detail them one by one.

\subsubsection{Lifting process.}
Given a 1D temporal signal \textit{x} = [$x_1,x_2, x_3,\dots, x_t $] ($x\in \mathcal{R} ^{C \times T},t \in \mathcal{N}+$) where $C$ denotes channel and $T$ represents the sequence length, lifting process decomposes \textit{x} into a downsized approximation \textit{s} and a difference signal \textit{d} as :
\begin{equation}
\label{e1}
s,d=\mathcal{F}(x).
\end{equation}   
Here, $\mathcal{F}(\cdot)$ = $f_{update}\circ  f_{predict}\circ f_{split}$ is consisted of three functions: split, predict and update as shown in Fig.~\ref{fig3}, where $\circ$ is the function composition operator.

\textbf{Split} $f_{split}$: $x \mapsto (x_e, x_o)$. It partitions input signal \textit{x} into two disjoint sets $x_e$, $x_o$ for downsized signal generation. Practically, $x_e$ and $x_o$ are generated with even and odd indices, respectively, where $x_e$= [$x_2,x_4,\dots, x_{2k} $] and $x_o$= [$x_1,x_3,\dots, x_{2k-1} $] ($k \in \mathcal{N}+$) are temporally closely correlated.

\textbf{Predict} $f_{predict}$: $(x_e, x_o) \mapsto d$. Given a selected set, e.g., $x_e$, $f_{predict}$ predicts another set $x_o$ by a predictor $\mathcal{P}(\cdot) $. As only one basis $x_e$ is given, the prediction is not required to be precise, which is expected to be collaborated with following $f_{update}$. So the difference signal \textit{d} with high-pass coefficients is obtained as :
\begin{equation}
\label{e2}
d=x_o-\mathcal{P}(x_e).
\end{equation}  

\textbf{Update} $f_{update}$: $(x_e, d)\mapsto s$. As the prediction process is undoubtedly aliasing and simply taking alternately sampled $x_e$ as the approximation of $x$ will cause inevitably information loss, an update function $\mathcal{U}(\cdot) $ takes the difference \textit{d} as input for compensation and generates the smoothed downsized representations \textit{s} as :
\begin{equation}
\label{e3}
s=x_e+\mathcal{U}(d).
\end{equation}  

This update procedure can be viewed as applying a low-pass filter for \textit{x} and thus \textit{s} is the downsized approximation of original signal with low-pass coefficients. It's worth noting that the prediction and update procedure are intrinsically adversarial. If $f_{predict}$ precisely predicts $x_o$ based on $x_e$, the difference signal \textit{d} will be minor and thus the approximation \textit{s} will be extremely biased towards $x_e$, resulting in aliasing downsampling. If $f_{predict}$ can't well perform prediction, the difference signal \textit{d} will be huge and make the approximation \textit{s} deformed from original signal \textit{x}. Thus, $f_{predict}$ and $f_{predict}$ works in an antagonistic way, expecting to generate discriminative and detailed signals, \textit{s} and \textit{d}, respectively.

The classic Lifting Scheme methods apply predefined low-pass filters and high-pass filters to decompose \textit{x} into different sub-bands. However, manually designing filters for $\mathcal{P}(\cdot) $ and $\mathcal{U}(\cdot)$ is difficult~\cite{zheng2010nonlinear} and can't fit various conditions. Previously, ~\cite{zheng2010nonlinear} proposed to optimize filters in $\mathcal{P}(\cdot) $ and $\mathcal{U}(\cdot)$ by backward propagated gradients for signal processing. We design $\mathcal{P}(\cdot) $ and $\mathcal{U}(\cdot)$ with tiny fully convolutional networks which are optimized in an end-to-end manner as:
\begin{equation}
\label{e4}
\mathcal{P}(\cdot)  = {\rm Tanh}() \circ {\rm Conv}(k=1) \circ {\rm ReLU}() \circ {\rm Conv}(k=K, g=C_{in}),   
\end{equation}  
\begin{equation}
\label{e5}
\mathcal{U}(\cdot)  = {\rm Tanh}() \circ {\rm Conv}(k=1) \circ {\rm ReLU}() \circ {\rm Conv}(k=K, g=C_{in}).   
\end{equation} 
Here \textit{k} denotes the kernel size and \textit{g} represent the group number for convolution. We prefer to first deploy a 1D depth-wise convolution with kernel size \textit{K} to aggregate local temporal patterns, followed by a ReLU activation. And then we use a normal 1$\times$1 convolution to enable channel-wise aggregation, followed by a Tanh activation for feature prediction. 

For generating discriminative representations, two loss constraints are employed apart from original task loss. Recall that the downsized approximation \textit{s} is originated from $x_e$ by Eq.~\ref{e3}, it's essentially close to $x_e$. We employ loss $C_u$ to encourage $x_e$ to approximate $x_o$ as well, by minimizing the L2-Norm distance between \textit{s} and $x_o$ as:
\begin{equation}
\label{e6}
C_u = \| s-x_o\|^2  = \| \mathcal{U}(d)+ x_e - x_o \|^2.
\end{equation}  

Another loss $C_p$ is used to minimizing the L2-Norm of difference signal \textit{d} as:
\begin{equation}
\label{e7}
C_p = \| d \|^2  = \| x_o - \mathcal{P}(x_e) \|^2.
\end{equation}  

Thus, for a certain task, the final loss functions can be written as:
\begin{equation}
\label{e8}
\mathcal{L}_{total} = \mathcal{L}_{task}+ \alpha_{u} C_u + \alpha_{p}C_p
\end{equation}  
where $\mathcal{L}_{task}$ is the loss for a certain temporal task, e.g., CSLR in this paper, and $\alpha_{u}$ and $\alpha_{p}$ are coefficients for $C_u$ and $C_p$, respectively. 

\subsubsection{Component weighting.}
The approximation \textit{s} and difference signal \textit{d} represent low-pass coefficients and high-pass coefficients for original signal \textit{x}, respectively, which can be viewed as dominating movement patterns and adhering action details for input sequences. As not all frequency components in \textit{s} or \textit{d} play an important role in depicting human dynamics, we design a component weighting module $f_{weight}$ to dynamically emphasize or suppress certain components in \textit{s} or \textit{d} for robust temporal representations.

Specially, for each timestamp \textit{t}, $f_{weight}$ aims to generate a specific coefficient for each channel, resulting in a total weight matrix $W \in \mathcal{R} ^{C\times T}$. We instantiate $f_{weight}$ with a tiny fully convolutional network followed by a Sigmoid function at the end, which is optimized in an end-to-end manner to dynamically decide the weights. Each value ranging from (0,1) in $W$ represents the importance of a certain component generated by the lifting process. Instead of directly multiplying $W$ with inputs for weighting, which we found badly hurts original representations, we perform component weighting in a residual way as:
\begin{equation}
\label{e9}
X^{out} = (W -\frac{1}{2}\mathds{1})\times X_{in} + X_{in}
\end{equation} 
where $\mathds{1}$ $\in \mathcal{R} ^{C\times T}$ is a full-one matrix. We first change values in $W$ from (0,1) into (-$\frac{1}{2}$, $\frac{1}{2}$) and multiple $W$ with $X_{in}$ for obtaining biased components. Adding the biased components with $X_{in}$ can effectively strengthen or weaken $X_{in}$, without hurting its original expressions.
\subsubsection{Fusion.}
Given the low-pass and high-pass coefficients $s^{*}$ and $d^{*}$ after component weighting, we devise three simple strategies to fuse them into a single and robust representation as the temporal downsized output.

\textbf{Sum.}  $s^{*}$ and $d^{*}$ are simply summed to combine their components as:
\begin{equation}
\label{e10}
y = s^* + d^*
\end{equation} 

\textbf{Concatenation.}  $s^{*}$ and $d^{*}$ are concatenated along channel dimension as:
\begin{equation}
\label{e11}
y = {\rm Concat}(s^*, d^*),
\end{equation} 
resulting in $y \in \mathcal{R} ^{2C\times T}$ with double capacity.

\textbf{Convolutional bottleneck.} We employ a tiny convolutional network consisted of sequences of convolution with kernel size 1, BatchNorm and ReLU to combine $s^{*}$ and $d^{*}$ as:
\begin{equation}
\label{e12}
y = {\rm ReLU(BN(Conv( Concat}(s^*, d^*)))).
\end{equation} 

\subsubsection{Discussion.} Max pooling and average pooling are two widely used pool methods in various tasks. However, they follow predefined mechanisms to select values, which may not be optimal. For example, max pooling typically puts all attention on the element with the largest activation. However, the assumption that the maximum activation stands for the most discriminative element, may not always be true. Besides, the max operator hinders gradient-based optimization where only the largest activation in a sampling region is assigned back-propagated gradients, which may slow down convergence. Although average pooling ensures all elements can contribute to outputs, it treats them equally which usually results in smoothed outputs and hurts small but discriminative responses. In this paper, we refer to Lifting Scheme from signal processing to decompose signals into different sub-bands with major movements or discriminative details, which naturally fit the problem. Our method dynamically generates the downsized output for each sample, which jumps out of the hand-crafted scope like max or average pooling. Besides, our method can smoothly behave like max or average pooling, which falls between them but keeps more representative features than both due to hierarchical signal decomposition.    

\section{Experiments}
\subsection{Datasets}
We evaluate our method on two commonly used large-scale datasets: RWTH-PHOENIX-Weather-2014 (PHOENIX14) and RWTH-PHOENIX-2014-Weather-T (PHOENIX14-T).

PHOENIX14~\cite{koller2015continuous} is recorded from the German TV weather by nine signers wearing dark clothes in front of a clean background. It contains 6841 sentences with a vocabulary of 1295 signs, divided into 5672 training instances, 540 development (Dev) instances and 629 testing (Test) instances. All videos are shot by 25 fps with resolution 210$\times$260.

PHOENIX14-T~\cite{camgoz2018neural} is available for both CSLR and Sign Language Translation (SLT) tasks which can be viewed as an extension of PHOENIX14. It contains 8247 sentences with a vocabulary of 1085 signs, split into 7096 training samples, 519 development (Dev) samples
and 642 testing (Test) samples.
\subsection{Training Details}
ResNet18~\cite{he2016deep} is adopted as the 2D CNN backbone for fair comparison with recent methods. The 1D CNN is consisted of a sequence of \{K5, P2, K5, P2\} layers where K$\sigma$ and P$\sigma$ denotes a 1D convolutional layer and a pooling layer with kernel size of $\sigma$, respectively. A two-layer BiLSTM with hidden size 1024 is adopted for long-term temporal modeling, followed by a fully connected layer for sentence prediction. We train our models for 80 epochs with initial learning rate 0.001 which is divided by 5 at epoch 40 and 60. Adam~\cite{kingma2014adam} optimizer is adopted as default with weight decay 0.001 and batch size 2. All input frames are first resized to 256$\times$256, and then randomly cropped to 224$\times$224 with 50\% horizontal flipping and 20\% temporal rescaling during training. During inference, a 224$\times$224 center crop is simply adopted. Following VAC~\cite{Min_2021_ICCV}, we employ the visual enhancement loss and visual alignment loss for additional visual supervision, with weights of 1.0 and 25.0, respectively. We only substitute two pooling layers in the 1D CNN with our TLP, as shown in Fig.~\ref{fig2}. The coefficients $\alpha_{u}$ and $\alpha_{p}$ for loss $C_u$ and $C_p$ are set as 0.001.

Word Error Rate (WER) is used as the metric of measuring similarity between predicted sentence and reference sentence. It's defined as the minimal number of substitution, insertion and deletion operations
to convert the predicted sentence to the reference sentence as:
\begin{equation}
\label{e13}
\rm WER = \frac{\#substitutions+\#insertions+\#deletions}{\#reference}.
\end{equation}
Note that the lower WER, the better.

\subsection{Ablation Study}

\setlength\tabcolsep{3pt}
\begin{table}[t]
\centering
\begin{subtable}[t]{\textwidth}
\centering
\begin{tabular}{c|c|c}
\hline
Configurations for $\mathcal{P}(\cdot)$ \& $\mathcal{U}(\cdot)$ & Dev(\%) & Test(\%)\\
\hline
\textit{K}=3  &  20.0 &  21.1\\
\textit{K}=4  &  19.9 &  21.1 \\
\textit{K}=5  & \textbf{19.7} & \textbf{20.8} \\
\textit{K}=7  & 19.9  & 21.0 \\
$Tanh \circ Conv(k=5) $ & 20.2 & 21.3\\
\hline
\end{tabular}
\caption{Effects of different configurations for $\mathcal{P}(\cdot)$ \& $\mathcal{U}(\cdot)$}
\label{subtab1}
\end{subtable}

\begin{subtable}[t]{\textwidth}
\centering
\begin{tabular}{c|c|c}
\hline
Component weighting & Dev(\%) & Test(\%)\\
\hline
$Sigmoid \circ IN\circ Conv(k=1) \circ Conv(k=5, g=C_{in}) $ & 20.1& 21.2\\
$Sigmoid \circ IN\circ Conv(k=\textbf{3})$ & 20.0 & 21.2 \\
$Sigmoid \circ IN\circ Conv(k=\textbf{5})$ & \textbf{19.7} & \textbf{20.8}\\
$Sigmoid \circ IN\circ Conv(k=\textbf{7})$ &  19.9 & 20.9 \\
$Sigmoid \circ \textbf{BN}\circ Conv(k=5)$ & 21.1 &  22.3\\
\hline
$X^{out} = W \times X_{in}$  &  20.9 & 21.9 \\
$X^{out} = (W -\frac{1}{2}\mathds{1})\times X_{in} + X_{in}$ & \textbf{19.7} & \textbf{20.8}\\
\hline
- & 20.2 & 21.4\\
Shared for \textit{s} \& \textit{d} & 19.9  & 20.9\\
Independent for \textit{s} \& \textit{d} & \textbf{19.7}& \textbf{20.8}\\
\hline
\end{tabular}
\caption{Effects of different configurations for component weighting.}
\label{subtab2}
\end{subtable}

\begin{subtable}[t]{0.53\textwidth}
\centering
\begin{tabular}{c|c|c}
\hline
Fusion & Dev(\%) & Test(\%)\\
\hline
Only $s^*$ & 20.2 & 21.2\\
Sum & \textbf{19.7} & \textbf{20.8} \\
Concatenation & 19.9 & 21.1\\
Convolutional bottleneck&  20.1 & 21.4\\
\hline
\end{tabular}
\caption{Effects of different fusion methods.}
\label{subtab3}
\end{subtable}
\begin{subtable}[t]{0.46\textwidth}
\centering
\begin{tabular}{c|c|c}
\hline
Locations for TLP & Dev(\%) & Test(\%)\\
\hline
-& 21.2 & 22.3 \\
First location & 20.1& 21.3\\
Second location & 20.3& 21.1\\
Two locations & \textbf{19.7} & \textbf{20.8}\\
\hline
\end{tabular}
\caption{Effects of locations for TLP}
\label{subtab4}
\end{subtable}
\caption{Ablation study for different modules of TLP on PHOENIX14 dataset.}
\label{tab1}
\end{table}

\subsubsection{Configurations for $\mathcal{P}(\cdot)$ \& $\mathcal{U}(\cdot)$.} Tab.~\ref{subtab1} ablates the performance when varying the kernel size \textit{K} for $\mathcal{P}(\cdot)$ \& $\mathcal{U}(\cdot)$. Notably, a larger kernel with more local aggregation ability consistently brings better performance. When \textit{K} reaches 7, it brings no more performance gain. We thus set \textit{K} as 5 by default. We then test other instantiations for $\mathcal{P}(\cdot)$ \& $\mathcal{U}(\cdot)$, e.g., a simple combination of Conv and Tanh, and found it obtains lower performance than current design.

\subsubsection{Configurations for component weighting.} In the upper part of Tab.~\ref{subtab2}, we first test different implementations for $f_{weight}$ to dynamically strengthen or weaken various components. Compared to the two-Conv counterpart in the top, we observe a simpler design, i.e. $Sigmoid \circ IN\circ Conv$, achieves better performance which we employ as default. When varying the kernel size for $f_{weight}$, \textit{k}=5 performs best among all candidates. We further compare the effect of normalization methods in $f_{weight}$ which are typically employed to accelerate convergence and promote performance. InstanceNorm(IN)~\cite{ulyanov2016instance} and commonly used BatchNorm(BN)~\cite{ioffe2015batch} are compared. We find that IN achieves more stable and superior performance than BN, which results from cross-batch normalization hurting the feature distribution for each sample. We then compare the choice of directly weighting with our residual architecture for component weighting. Seen from the middle in Tab.~\ref{subtab2}, the residual architecture outperforms directly weighting by a large margin, where the latter inevitably hurts the original representation and leads to unstable expressions. We finally test the effects of component weighting under different configurations. Compared to w/o component weighting, deploying component weighting in an either shared or independent way for \textit{s} \& \textit{d} achieves better performance. Furthermore, independent weighting for \textit{s} \& \textit{d} brings more performance boost by considering the specific characteristics of two pathways. 

\subsubsection{Fusion methods.} Tab.~\ref{subtab3} ablates different configurations for fusion of $s^*$ and $d^*$. $s^*$ and $d^*$ generated by lifting process correspond to different hierarchical features, while our TLP allows to flexibly choose which sub-band to be kept as downsized outputs. We note that only preserving $s^*$ obtains lower performance than other variants, which demonstrates the effectiveness of combining low-pass coefficients $s^*$ and high-pass coefficients $d^*$ for effective recognition. Tab.~\ref{subtab3} shows that simply summing $s^*$ and $d^*$ gives the best performance among all candidates, which we employ as default in the following experiments.

\subsubsection{Locations for TLP.} We incrementally add one or more TLPs in different locations to verify its effectiveness in Tab.~\ref{subtab4}. Compared to our baseline w/o TLP, adding one TLP in either the first or second location brings considerable 1.1\% and 0.9\% performance boost, respectively. Replacing total two pooling instances with TLP gives notable 1.5\% promotion without any other architecture change, demonstrating the key role of temporal pooling and the effect of TLP for robust discriminative representations.

\begin{table}
\centering
\setlength\tabcolsep{2pt}
\begin{tabular}{cccccc}
\hline
Methods & GFLOPs & Throughout(Vids/s) & Memory(M) & Dev(\%) & Test(\%)\\ 
\hline
Max pooling   & 3.671 & 12.22 & 8827& 21.2 & 22.3\\
Average pooling & 3.671 & 12.40 & 8827 & 21.1 & 22.1\\
TLP(Ours) &3.686 & 12.12& 8846& 19.7 (+\textbf{1.5}) &20.8 (+\textbf{1.5})\\
\hline
\end{tabular}  
\caption{Computational efficiency of TLP against commonly used max pooling and average pooling on PHOENIX14 dataset.}
\label{tab2}  
\end{table}

\subsection{Computational Efficiency}
Our TLP is an efficient plug-and-play tool with little computational overhead. Under the formula of Eq.~\ref{e4}, Eq.~\ref{e5} and Eq.~\ref{e9}, two TLPs totally consumes 7.5M$\times$2 = 15.0M FLOPs\footnote{FLOPs denote floating point operations, which measure the computational costs of models.}, which is negligible (0.4\%) compared to our 2D backbone ResNet18 (3.64G FLOPs). Considering our TLP is composed of highly specialized operators like convolution and activation functions, it enjoys high computational efficiency on GPUs with little delay. As shown in Tab.~\ref{tab2}, compared to commonly used max pooling and average pooling, our TLP exhibits similar GFLOPs, throughout and memory usage with significantly higher performance, which is a both effective and efficient plug-and-play tool for temporal tasks.

\begin{table}[t]   
\centering
\begin{tabular}{ccccc}
\hline
\multirow{2}{*}{Pooling methods} &\multicolumn{2}{c}{PHOENIX14} & \multicolumn{2}{c}{PHOENIX14-T} \\
&  Dev(\%) & Test(\%)  & Dev(\%) & Test(\%)\\
\hline
Max pooling (\textbf{Baseline}) & 21.2 & 22.3 & 21.1 & 22.8\\
Stochastic pooling & 22.2 (-1.0) & 23.4 (-1.1)& 22.3 (-1.2) & 23.7 (-0.9)\\
Mixed pooling & 21.5 (-0.3) & 22.6 (-0.3)& 21.3 (-0.2) & 23.0 (-0.2)\\
$L_p$ pooling ($p$=3) & 21.5 (-0.3) & 22.5 (-0.2)& 21.3 (-0.2) & 23.1 (-0.3)\\
SoftPool & 21.3 (-0.1) & 22.5 (-0.2)& 21.1 (+0.0) & 22.9 (-0.1)\\
$L_p$ pooling ($p$=2) & 21.1 (+0.1) & 22.3 (+0.0)& 21.2 (-0.1) & 22.7 (+0.1)\\
Average pooling & 21.1 (+0.1) & 22.1 (+0.2)& 20.9 (+0.2) & 22.6 (+0.2)\\
\hline
TLP(Ours) &19.7 (\textbf{+1.5}) & 20.8 (\textbf{+1.5}) & 19.4 (\textbf{+1.7}) & 21.2 (\textbf{+1.6})\\
\hline
\end{tabular}    
\caption{Comparison of our TSP with other pooling variants on the Dev and Test set of PHOENIX14 and PHOENIX14-T dataset. } 
\label{tab3}
\end{table}

\subsection{Comparison with other pooling methods}
We compare our TSP with other pooling variants to demonstrate
its effectiveness in Tab.~\ref{tab3}. Most of these counterpart pooling methods are elaborately designed for preserving critical spatial features. As shown in Tab.~\ref{tab3}, except $L_p$ pooling (\textit{p}=2) and average pooling, most pooling methods cause performance decline on both PHOENIX14 dataset and PHOENIX14-T dataset. $L_p$ pooling (\textit{p}=2) and average pooling bring marginal performance boost ($\leq$ 0.2\%). Although most of these variants are elaborately designed for spatial tasks with superior performance, they don't exhibit much superiority on temporal tasks, e.g., CSLR. In contrast, some of them even lead to lower performance. Hand-crafted pooling methods, like max pooling and average pooling, show robust performance on both datasets, demonstrating their excellent generalization ability. Compared to these pooling variants, our TLP consistently exhibits superior performance ($\geq$ 1.5\%) upon both datasets and largely surpasses all of them by a large margin. These results verify the effectiveness of our TLP by combining different sub-bands for discriminative hierarchical representations.

\begin{table}
\centering
\begin{tabular}{lllll}
\hline
\makecell[c]{\multirow{2}{*}{Methods}} &\multicolumn{2}{c}{PHOENIX14} & \multicolumn{2}{c}{PHOENIX14-T} \\
&  Dev(\%) & Test(\%)  &  Dev(\%) & Test(\%)  \\
\hline
ResNet18~\cite{he2016deep}  & 21.2 & 22.3 & 21.1 & 22.8 \\
ResNet18 w/ TLP &19.7 (\textbf{+1.5}) & 20.8 (\textbf{+1.5}) & 19.4 (\textbf{+1.7}) & 21.2 (\textbf{+1.6})\\
\hline
SqueezeNet~\cite{iandola2016squeezenet}  & 23.2 & 23.5 & 21.7 & 23.1 \\
SqueezeNet w/ TLP & 22.2 (\textbf{+1.0}) & 22.3 (\textbf{+1.2}) & 20.6 (\textbf{+1.1})& 22.0 (\textbf{+1.1})\\
\hline
RegNetX-800Mf~\cite{radosavovic2020designing}  & 21.4 & 22.5 & 21.0 & 22.3 \\
RegNetX-800Mf w/ TLP & 20.0 (\textbf{+1.4}) & 21.1 (\textbf{+1.4}) & 19.5 (\textbf{+1.5}) & 20.9 (\textbf{+1.4})\\
\hline
RegNetY-800Mf~\cite{radosavovic2020designing}  & 21.3 & 22.2 & 20.7 & 21.8 \\
RegNetY-800Mf w/ TLP & 19.7 (\textbf{+1.6})& 20.5 (\textbf{+1.7})& 19.1 (\textbf{+1.6})& 20.1 (\textbf{+1.7}) \\
\hline
\end{tabular}  
\caption{Generalizability of TLP upon various backbones on both PHOENIX14 dataset and PHOENIX14-T dataset.}
\label{tab4}  
\end{table}

\subsection{Generalizability}
We apply TLP to several backbones including ResNet18~\cite{he2016deep}, SqueezeNet~\cite{iandola2016squeezenet}, RegNetX-800Mf~\cite{radosavovic2020designing} and RegNetY-800Mf~\cite{radosavovic2020designing} on both datasets to demonstrate its generalizability. As shown in Tab.~\ref{tab4}, TLP consistently brings significant performance boost ($\geq$ 1.0\%) across different backbones. We observe an interesting phenomenon where the effect of TLP seems to be proportional to the strength of backbones. For example, the boost by TLP is relatively smaller (1.0\%) for less powerful SqueezeNet~\cite{iandola2016squeezenet} (23.2\%). In contrast, TLP brings much more performance boost ($\geq$ 1.7\%) for more powerful backbones, e.g., RegNetY-800Mf~\cite{radosavovic2020designing} (21.3\%). Other backbones like MobileNet-V2~\cite{howard2018inverted}, EfficientNet-B0~\cite{tan2019efficientnet} and MNASNet~\cite{tan2019mnasnet} are found out of memory upon current devices.

\begin{table}[t]   
\centering
\begin{tabular}{cccccccc}
\hline
\multirow{3}{*}{Methods} &\multirow{3}{*}{Backbone}  & \multicolumn{4}{c}{PHOENIX14} & \multicolumn{2}{c}{PHOENIX14-T} \\
& & \multicolumn{2}{c}{Dev(\%)} & \multicolumn{2}{c}{Test(\%)} &  \multirow{2}{*}{Dev(\%)} & \multirow{2}{*}{Test(\%)}\\
& & del/ins & WER & del/ins& WER & & \\
\hline
SubUNet~\cite{cihan2017subunets}& CaffeNet &14.6/4.0 & 40.8 &4.3/4.0 &40.7 &- &-\\
Staged-Opt~\cite{cui2017recurrent}& VGG-S  &13.7/7.3& 39.4 &12.2/7.5 & 38.7 &- &-\\
Align-iOpt~\cite{pu2019iterative}& 3D-ResNet&12.6/2 & 37.1& 13.0/2.5 & 36.7 & -&-\\
Re-Sign~\cite{koller2017re}& GoogLeNet & - & 27.1 &- &26.8 &- &-\\
SFL~\cite{niu2020stochastic}& ResNet18& 7.9/6.5 & 26.2 & 7.5/6.3& 26.8 & 25.1&26.1\\
STMC~\cite{zhou2020spatial}& VGG11 &  - & 25.0 &  - & - & -& -\\
DNF~\cite{cui2019deep}& GoogLeNet  & 7.8/3.5 & 23.8 & 7.8/3.4 & 24.4 & -&-\\
FCN~\cite{cheng2020fully}& Custom & - & 23.7 & -& 23.9 & 23.3& 25.1\\
CMA~\cite{pu2020boosting} & GoogLeNet & 7.3/2.7 &\textbf{21.3} & 7.3/2.4 & \textbf{21.9}  & -&-\\
VAC~\cite{Min_2021_ICCV}& ResNet18 & 7.9/2.5 & \textbf{21.2} &8.4/2.6 & \textbf{22.3} &- &-\\
\hline
SLT$^*$~\cite{camgoz2018neural}& GoogLeNet  & - & - & - & - & 24.5 & 24.6\\
C+L+H$^*$~\cite{koller2019weakly}& GoogLeNet  & - &26.0 & - & 26.0 & 22.1 & 24.1 \\
DNF$^*$~\cite{cui2019deep}& GoogLeNet  & 7.3/3.3 &23.1& 6.7/3.3 & 22.9 & - & -\\
STMC$^*$~\cite{zhou2020spatial}& VGG11 & 7.7/3.4 &\textbf{21.1} & 7.4/2.6 & \textbf{20.7} & \textbf{19.6} & \textbf{21.0}\\
\hline
Baseline & ResNet18 & 7.9/2.5 & 21.2 &8.4/2.6 & 22.3 & 21.1 & 22.8\\
\textbf{TLP(Ours)} & ResNet18 & 6.3/2.8  &\textbf{19.7} &   6.1/2.9& \textbf{20.8}  & \textbf{19.4} & \textbf{21.2} \\
\hline   
\end{tabular}  
\caption{Comparison with other state-of-the-art methods on the PHOENIX14 and PHOENIX14-T datasets. $*$ indicate extra clues such as face or hand features are included. 'C+L+H' denotes the abbreviation of 'CNN+HMM+LSTM~\cite{koller2019weakly}'} 
\label{tab5}
\end{table}

\subsection{Comparison with the state-of-the-art}
We compare our model against other state-of-the-art methods
on the PHOENIX14 and PHOENIX14-T dataset in Tab.~\ref{tab5}. The entries notated with $*$ indicate these methods such as SLT~\cite{camgoz2018neural}, CNN+LSTM+HMM~\cite{koller2019weakly}, DNF~\cite{cui2019deep} and STMC~\cite{zhou2020spatial} utilize additional factors like face or hand features for better performance. We observe that our method outperforms all previous approaches with video information only. We also notice that our method even surpasses those approaches equipped with additional factors when only using video information, which demonstrates the effectiveness of our TLP.

\begin{figure}[t]
\centering
\includegraphics[width=\columnwidth]{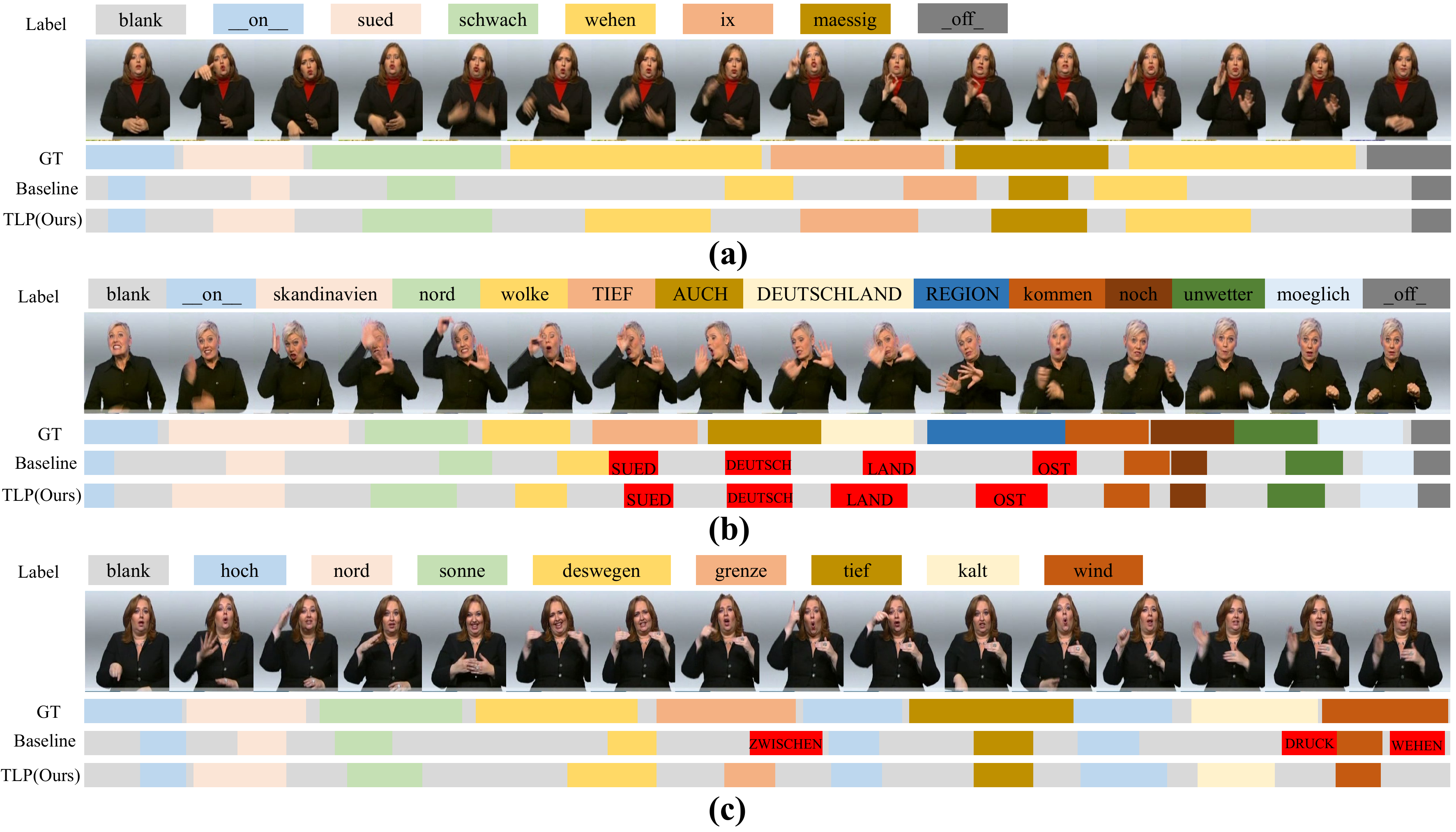} 
\caption{Predictions of our baseline and TLP for several example videos. (a) All glosses are correctly recognized by both our baseline and TLP. (b) The same four glosses are wrongly recognized by both our baseline and TLP. (c) Our baseline wrongly recognizes several glosses  while TLP makes correct predictions. Wrong recognized glosses (except del) are marked in red.}
\label{fig4}
\end{figure}
 
\section{Visualizations}
To better understand the effects of TLP, we visualize several videos with their predictions of our baseline and TLP from the Dev set of PHOENIX14 dataset in Fig.~\ref{fig4}. Wrong recognized glosses (except del) are marked in red. We show three different cases to help demonstrate the effect of TLP in various conditions. All glosses in Fig.~\ref{fig4}(a) are correctly recognized by both our baseline and TLP. The same four glosses in Fig.~\ref{fig4}(b) are wrongly recognized by both our baseline and TLP. Our baseline wrongly recognizes several glosses in Fig.~\ref{fig4}(c) while TLP makes correct predictions. We notice in all cases, TLP predicts more centralized gloss borders than our baseline which helps accurate recognition.

\section{Conclusion}
In this paper, we derive temporal lift pooling (TLP) from the Lifting Scheme in signal processing to decompose input signals into various sub-bands, each corresponding to a specific movement pattern. Combining different components of TLP results in a refined downsized feature map, well preserving discriminative representations. TLP exhibits excellent generalizability upon multiple backbones upon two large-scale CSLR datasets with significant performance boost. Visualizations verify the effects of TLP for correcting gloss borders.

\textbf{Acknowledgment.} This work is supported by NSFC 62072334 Project.
%
%
\bibliographystyle{splncs04}
\bibliography{ref}

\appendix
\title{Supplementary Material for Temporal Lift Pooling for Continuous Sign Language Recognition} 
\titlerunning{Supplementary Material for Temporal Lift Pooling for CSLR}

\author{Lianyu Hu\inst{1}\orcidlink{0000-0003-2470-8110} \and
Liqing Gao\inst{1}\orcidlink{0000-0003-4518-2154} \and
Zekang Liu\inst{1}\orcidlink{0000-0002-5003-1900}\and
Wei Feng\textsuperscript{\Letter}\inst{1}\orcidlink{0000-0003-3809-1086}}

\authorrunning{Lianyu et al.}

\institute{Tianjin University, Tianjin 300350, China
\email{\{hly2021,lqgao,lzk100953\}@tju.edu.cn;wfeng@ieee.org}}
\maketitle
\section{Ablations of Loss $C_u$ and $C_p$}
\textbf{Ablations of Loss $C_u$ and $C_p$ } are given in tab.~\ref{tab1}. One can notice that employing either loss only give little accuracy boost. Employing both losses benefits most.
\begin{table}[h]   
   \centering
   \begin{tabular}{cccccc}
   \hline
   \multirow{2}{*}{$C_u$} & \multirow{2}{*}{$C_p$} &\multicolumn{2}{c}{PHOENIX14} & \multicolumn{2}{c}{PHOENIX14-T}\\
   & &  Dev(\%) & Test(\%)  &  Dev(\%) & Test(\%)\\
   \hline
   \ding{53} & \ding{53} & 20.8 & 21.7  & 20.4 & 22.0 \\
   \ding{53} & $\checkmark$ & 20.4 & 21.4  & 19.9 & 21.7 \\
   $\checkmark$ & \ding{53} & 20.3 & 21.3  & 20.0 & 21.7 \\
   $\checkmark$ & $\checkmark$ &\textbf{19.7} & \textbf{20.8}  & \textbf{19.4} & \textbf{21.2} \\
   \hline
   \end{tabular}    
\caption{Effectiveness of Loss $C_u$ and $C_p$ on two datasets.} 
\label{tab1}  
\end{table}

\section{Flexibility of TLP upon other CSLR methods}
Tab.~\ref{tab3} shows the results of equipping FCN~\cite{cheng2020fully} with TLP, which gains +1.8\% WER boost on average. As no official code is released for FCN, we manually reproduce it.
\begin{table}[h]   
   \setlength\tabcolsep{2pt}
   \centering
   \begin{tabular}{ccccc}
   \hline
   \multirow{2}{*}{Methods} &\multicolumn{2}{c}{PHOENIX14} & \multicolumn{2}{c}{PHOENIX14-T}\\
   &  Dev(\%) & Test(\%)  &  Dev(\%) & Test(\%)\\
   \hline 
   FCN [3] & 24.2 & 24.5  & 23.6 & 25.2 \\
   FCN w/ TLP  &22.4(\textbf{+1.8}) & 22.7(\textbf{+1.8})   & 21.9(\textbf{+1.7})  & 23.1(\textbf{+2.1})  \\
   \hline
   \end{tabular}     
\caption{Flexibility of TLP upon FCN [3] on two datasets.} 
\label{tab3}
\end{table}

\section{Results on the CSL dataset}
Tab.~\ref{tab2} validates the effectiveness of our TLP on the chinese sign language dataset (CSL)~\cite{huang2018video}. The CSL dataset is collected within laboratory environments with a vocabulary size of 178 with 100 sign language sentences. Each sentence is performed by fifty signers with five times, resulting in total 25000 videos with 100+ hours. Our method achieves significant accuracy boost (+5.5\%) on the CSL dataset.

\begin{table}[h]   
   \centering
   \begin{tabular}{cc}
   \hline
   Methods &CSL~\cite{huang2018video}(\%)\\
   \hline
    SubUNet~\cite{cihan2017subunets} & 11.0\\
    FCN~\cite{cheng2020fully} & 3.0\\
    STMC~\cite{zhou2020spatial}* & 2.1\\
    VAC~\cite{Min_2021_ICCV} & 1.6 \\
    \hline
   Baseline & 7.3 \\
   Baseline w/ TLP & 1.8 (+\textbf{5.5}) \\
   \hline
   \end{tabular}     
\caption{Effectiveness of TLP on the CSL Dataset.} 
\label{tab2} 
\end{table}

%
%
\end{document}